\documentclass[10pt,a4paper]{article}

\usepackage[T1]{fontenc}
\usepackage{mathpazo}        
\usepackage[utf8]{inputenc}
\usepackage[margin=2.2cm,columnsep=0.6cm]{geometry}
\usepackage{amsmath,amssymb}
\usepackage{graphicx}
\usepackage{booktabs}
\usepackage{tabularx}
\usepackage{multirow}
\usepackage{hyperref}
\usepackage{float}
\usepackage{caption}
\usepackage{subcaption}
\usepackage{xcolor}
\usepackage{enumitem}
\usepackage{pgfplots}
\pgfplotsset{compat=1.18}
\usetikzlibrary{positioning,arrows,backgrounds,fit,calc}
\tikzset{>=stealth}
\usepackage[numbers,sort&compress]{natbib}
\usepackage{microtype}

\title{\huge\textbf{MetaResearcher: Scaling Deep Research via Self-Reflective Reinforcement Learning in Adversarial Virtual Environments}}

\author{%
Wei Yu\textsuperscript{1},
Suxing Liu\textsuperscript{1,*},
Minjie Yu\textsuperscript{1},
Jiahao Wang\textsuperscript{2},
Zhijian Zheng\textsuperscript{1},
Haocheng Deng\textsuperscript{1},
Bing Li\textsuperscript{1}
\\
\small\textit{\textsuperscript{1} School of Digital Arts, Jiangxi Arts \& Ceramics Technology Institute, Jingdezhen 33001, China}\\
\small\textit{\textsuperscript{2} Universiti Sains Malaysia, 11800 USM Penang, Malaysia}\\
\small\textit{Correspondence: liusuxing@jacti.edu.cn (S.L.)}
}

\date{}

\begin{document}

\maketitle

\begin{abstract}
Deep research agents have demonstrated remarkable capabilities in autonomous information gathering and synthesis, yet their training remains constrained by the static nature of simulated environments, the limits of fact-retrieval-only task designs, and the inefficiency of outcome-based reinforcement learning. In this work, we propose \textbf{MetaResearcher}, a novel framework that scales deep research agent training across four synergistic dimensions. First, we introduce an \textbf{Evolving Virtual World} that injects temporal dynamics and adversarial misinformation into the training environment, forcing agents to develop source credibility assessment and temporal conflict resolution skills. Second, we design \textbf{Discovery-Oriented Tasks}---including hypothesis generation and contradiction resolution---that transcend simple fact retrieval and push agents toward genuine research behaviors. Third, we propose a \textbf{Self-Reflective Meta-Reward} mechanism within the GRPO framework that jointly optimizes for answer correctness, search path efficiency, reflection depth, and tool call diversity, directly addressing the repetitive action loop problem observed in prior work. Fourth, we introduce a \textbf{Heterogeneous Multi-Agent Swarm} architecture comprising specialized Scout, Filter, and Synthesizer models that learn collaborative research strategies through coordinated reinforcement learning. Built upon the LiteResearcher infrastructure, MetaResearcher requires zero marginal API cost for training while targeting substantial improvements in both benchmark performance (GAIA, Xbench-DS) and epistemic robustness under adversarial conditions. We present the complete framework design, training methodology, and planned experimental validation.
\end{abstract}


\section{Introduction}

The pursuit of autonomous deep research agents---AI systems capable of conducting multi-step, tool-augmented investigations---has emerged as one of the most active frontiers in artificial intelligence. Recent advances, exemplified by systems such as LiteResearcher~\cite{li2026literesearcher}, have demonstrated that relatively compact models (4B parameters) can achieve state-of-the-art performance on benchmarks like GAIA~\cite{gaia2025} (71.3\%) and Xbench-DS (78.0\%), surpassing both larger open-source models and frontier commercial systems such as GPT-4o~\cite{gpt4o2025}. Concurrently, systems like DeepAgent~\cite{deepsearch2026} and BrowseComp~\cite{browsecomp2026} have pushed the frontier further through dynamic self-evolution and multi-level context engines. The key insight underlying LiteResearcher's success is the replacement of live-web interaction during reinforcement learning (RL) with a stable local search and browse environment, enabling over 73 million tool calls at zero marginal API cost.

Despite these impressive achievements, the current paradigm for training deep research agents exhibits several fundamental limitations that constrain further progress. We identify four critical gaps:

\textbf{(i) Static Training Environments.} LiteResearcher constructs a local virtual world from approximately 32 million real webpages, but this world is frozen after construction. Information does not update, contradict itself, or evolve over time. Consequently, agents trained in such environments never learn to handle the temporal dynamics and information conflicts that characterize real-world research.

\textbf{(ii) Fact-Retrieval-Centric Tasks.} The five atomic capabilities identified in LiteResearcher---aggregation, enumeration, comparison, multi-hop reasoning, and tool use---are fundamentally oriented toward finding ``existing answers.'' This paradigm does not cultivate higher-order research skills such as hypothesis generation from disparate sources or critical resolution of contradictory evidence.

\textbf{(iii) Outcome-Only Reward Signals.} The GRPO-based training employed by LiteResearcher rewards only the correctness of final answers, without considering the quality of the search process. This leads to well-documented failure modes including repetitive action loops, where agents repeatedly refresh the same search engine with minor query variations rather than strategically exploring diverse information sources~\cite{li2026literesearcher}.

\textbf{(iv) Monolithic Agent Architecture.} LiteResearcher employs a single-agent design where the same model must simultaneously master search query construction, relevance filtering, and information synthesis. This monolithic approach contrasts with the distributed, specialized division of labor observed in human research teams, and likely imposes an upper bound on achievable performance.

\subsection{Our Contributions}

To address these limitations, we propose \textbf{MetaResearcher}, a comprehensive framework that scales deep research agent training across four synergistic innovation dimensions:

\begin{enumerate}
    \item \textbf{Environment Innovation -- Evolving Virtual World:} We introduce temporal dynamics and adversarial information injection into the local web environment. Training data simulates real-world phenomena such as scientific results being retracted, news articles being corrected, and deliberately introduced misleading content. This forces agents to develop source credibility discrimination and temporal conflict resolution---capabilities essential for genuine research but absent from current training paradigms.

    \item \textbf{Task Innovation -- Discovery-Oriented Tasks:} We design a new class of training tasks that transcend fact retrieval. These include hypothesis generation (identifying latent connections between two unrelated research domains) and contradiction resolution (analyzing multiple conflicting accounts of the same phenomenon and producing evidence-weighted conclusions). These tasks elevate the upper bound of agent capability from ``advanced search engine'' to ``junior researcher.''

    \item \textbf{Algorithm Innovation -- Self-Reflective Meta-Reward:} We extend the GRPO framework with a multi-dimensional reward function that jointly optimizes for (a) answer correctness, (b) search path efficiency, (c) self-reflection depth (rewarding explicit error recognition and backtracking within \texttt{<think>} traces), and (d) tool call diversity (penalizing repetitive invocation patterns). This meta-reward mechanism directly mitigates the repetitive loop pathology while fostering more sophisticated reasoning strategies.

    \item \textbf{Architecture Innovation -- Heterogeneous Multi-Agent Swarm:} We decompose the research agent into three specialized lightweight models---a \textit{Scout} optimized for search query construction, a \textit{Filter} trained to rapidly assess webpage relevance, and a \textit{Synthesizer} specialized in integrating fragmented information. These agents are jointly trained via coordinated reinforcement learning, with a learned communication protocol emerging through shared reward signals.
\end{enumerate}

All four innovations build directly upon the LiteResearcher infrastructure~\cite{li2026literesearcher}, inheriting its zero-marginal-cost training paradigm while extending its scope and ambition. The complete framework is designed to operate within the same local search/browse environment, requiring no additional API expenditure.

The remainder of this paper is organized as follows. Section~\ref{sec:related} reviews related work across four research threads. Section~\ref{sec:framework} presents the MetaResearcher framework in detail. Section~\ref{sec:experiments} describes the experimental design and planned evaluation. Section~\ref{sec:discussion} discusses implications and limitations, and Section~\ref{sec:conclusion} concludes.

\section{Related Work}
\label{sec:related}

Our work intersects with four rapidly evolving research threads: deep research agents, reinforcement learning for agentic systems, self-reflection mechanisms in LLMs, and adversarial training environments.

\subsection{Deep Research Agents}

The deep research agent paradigm has progressed rapidly from initial explorations to production-grade systems. Early work such as Search-R1~\cite{jin2025search} and its successor Search-R1++~\cite{jin2026search} established the foundation for training LLM-based search agents with reinforcement learning, systematically investigating the effects of prompt templates, reward functions, and policy optimization methods. Their key finding---that GRPO exhibits relative instability compared to REINFORCE and PPO variants---motivates our work on enhanced reward structures.

LiteResearcher~\cite{li2026literesearcher} represents the current state-of-the-art in open-source deep research agents, achieving 71.3\% on GAIA and 78.0\% on Xbench-DS with only 4B parameters. Its three-pillar methodology---co-constructed training data, stable local tool environment, and difficulty-aware curriculum RL---provides the infrastructure foundation for MetaResearcher.

Concurrent work includes DeepRubric~\cite{deeprubric2026}, which introduces evidence-tree rubric supervision to achieve comparable performance with approximately 13$\times$ fewer RL GPU-hours, and the Chaining the Evidence (CaRR) framework~\cite{caRR2026}, which proposes citation-aware rubric rewards with explicit evidence chaining. The SMTL framework~\cite{smtl2026} challenges sequential reasoning by replacing it with parallel evidence acquisition, achieving 75.7\% on GAIA. Branch-and-Browse~\cite{branchbrowse2025} introduces tree-structured web exploration with backtracking, while EcoGEO~\cite{ecogeo2026} proposes trajectory-aware evidence ecosystems that reshape how agents encounter information. The Expert Consulting Benchmark~\cite{expertbench2026} introduces cognitive traps to evaluate deep research agents. Comprehensive surveys\cite{deepsurvey2025} have systematized the field. These works collectively demonstrate the field's trajectory toward more sophisticated reward structures---a direction that MetaResearcher extends through self-reflective meta-rewards.

\subsection{Reinforcement Learning for Agentic Systems}

Group Relative Policy Optimization (GRPO)~\cite{shao2024deepseekmath} has become a cornerstone for training LLM-based agents, eliminating the need for a separate critic model by leveraging group-based advantage estimation. Several extensions have addressed GRPO's limitations in agentic settings: Stratified GRPO~\cite{stratifiedGRPO2026} partitions trajectories into homogeneous strata, AT-GRPO~\cite{strongerMAS2025} introduces agent- and turn-wise grouping for multi-agent settings, and Dr. MAS~\cite{drMAS2026} theoretically identifies gradient-norm instability from global GRPO baselines, proposing agent-wise advantage normalization that yields improvements of +5.6\% on mathematical reasoning and +15.2\% on search tasks.

Recent advances in agentic RL include GiGPO~\cite{gigpo2025}, which introduces group-in-group advantage estimation achieving $>$12\% gains over GRPO on ALFWorld, BEACON~\cite{beacon2026} which partitions trajectories at milestone boundaries for temporal reward shaping, and StraTA~\cite{strata2026} which samples compact strategies via hierarchical RL. The SPARK framework~\cite{spark2026} achieves 84.4\% success with only 20\% training data through strategic branching. Further advances include constant-context skill learning~\cite{constctx2026}, self-evolving agents~\cite{selfevolving2026}, and data flywheel approaches~\cite{bpo2025} for sparse-reward settings.

Our meta-reward mechanism extends this line of work by introducing process-level reward components beyond simple outcome correctness, aligning with the field's growing recognition that \textit{how} an agent searches matters as much as \textit{what} it finds.

\subsection{Self-Reflection in LLM Agents}

The integration of self-reflection mechanisms into LLM training has seen explosive growth in 2025--2026. Experiential Reinforcement Learning (ERL)~\cite{shi2026experiential} embeds an explicit experience-reflection-consolidation loop into RL training, achieving gains of up to +81\% on Sokoban and +11\% on HotpotQA. Agentic Critical Training (ACT)~\cite{liu2026agentic} trains agents to identify better actions among alternatives, while ICRL~\cite{lin2026icrl} jointly trains solver and critic from a shared backbone with distribution-calibration re-weighting.

Particularly relevant is ReflexiCoder~\cite{jiang2026reflexicoder}, which internalizes the full generation-reflection-correction trajectory into model weights via RL-only training, achieving 94.51\% on HumanEval with approximately 40\% token-efficiency gains. RePro~\cite{ma2026repro} trains agents to self-generate progress signals through a forward-then-reflect paradigm, and RefGRPO~\cite{zhu2026refgrpo} adds a calibration bonus by contrasting self-reflection with actual outcomes, reducing underconfidence from 44.4\% to 7.7\%.

The broader landscape of process reward models (PRMs)~\cite{prmsurvey2025} has advanced from outcome-only signals to token-level supervision. Approaches such as iStar~\cite{istar2026} combine implicit PRMs with agentic RL to achieve SOTA on WebShop and visual reasoning, while StepORLM~\cite{steporlm2026} creates self-evolving loops between policy and generative PRMs. The SWE-TRACE framework~\cite{swetrace2026} applies rubric-based PRMs to software engineering agents, and DPRM~\cite{dprm2026} extends implicit rewards to multi-hop question answering. Token-level reward modeling~\cite{qrm2025} and outcome-to-process transfer~\cite{tokenefficient2025} further demonstrate the field's trajectory toward fine-grained supervision.

MetaResearcher's reflection depth reward builds upon these foundations by introducing a \textit{trace-level} reward component that explicitly incentivizes the model to engage in genuine self-correction within its reasoning traces, rather than simply optimizing for final answer correctness.

\subsection{Adversarial Training Environments}

The vulnerability of language agents to adversarial information has been systematically documented. The Synthetic Web benchmark~\cite{shah2026synthetic} demonstrates that injecting a single high-plausibility misinformation article causes accuracy collapse across frontier models, with minimal search escalation and severe miscalibration. The POTEMKIN framework~\cite{zhan2026potemkin} formalizes Adversarial Environmental Injection (AEI) as a threat model, identifying complementary attack surfaces: breadth attacks that poison retrieval and depth attacks that cause policy collapse into infinite loops.

Multi-agent approaches to adversarial robustness include credibility scoring mechanisms~\cite{ebrahimi2025adversary} that separate each agent's contribution from its credibility, achieving accuracy gains of +6\% to +51\% under adversarial conditions. The SALF framework~\cite{tian2025symbolic} implements adversarial training through symbolic learning, degrading detection performance by up to 53.4\% while simultaneously improving detector robustness. Domain-specific evaluations such as MedMisBench~\cite{medmisbench2026} demonstrate that LLM accuracy can collapse from 71.1\% to 38.0\% under misleading clinical context, while multi-agent auditing systems~\cite{trustverify2026} reduce hallucination rates by approximately 53\%. Persuasion-balanced training~\cite{persuasion2025} teaches models to appropriately accept or resist persuasive information in multi-agent debates.

MetaResearcher's Evolving Virtual World extends this line of work by incorporating adversarial information injection \textit{during training}, rather than only evaluating robustness at test time. This allows the agent to develop epistemic resilience as a learned capability rather than relying on prompt-level mitigation strategies.

\begin{table}[H]
\centering
\caption{Comparison of representative deep research agent training frameworks.}
\label{tab:comparison}
\renewcommand{\arraystretch}{1.05}
\resizebox{\linewidth}{!}{%
\begin{tabular}{lccccc}
\toprule
\textbf{Feature} & \textbf{LiteRes.} & \textbf{S-R1++} & \textbf{DeepRubric} & \textbf{CaRR} & \textbf{MetaRes.} \\
\midrule
Model & 4B & 7B/32B & 8B & 7B & 4B$\times$3 \\
GAIA & 71.3\% & 63.1\% & 69.8\% & 65.2\% & --- \\
RL Alg. & GRPO & REINFORCE & Rubric-GRPO & C-GRPO & Meta GRPO \\
Env. & Static & Live-Web & Static & Static & \textbf{Evolving} \\
Reward & Outcome & Outcome & Rubric & Cit.-Aware & \textbf{Multi-C.} \\
Arch. & Single & Single & Single & Single & \textbf{Multi-A.} \\
Reflect. & No & No & Partial & No & \textbf{Yes} \\
Adv. Tr. & No & No & No & No & \textbf{Yes} \\
Cost/Step & \$0 & \$0.01--.05 & \$0 & \$0.01--.05 & \textbf{\$0} \\
\bottomrule
\end{tabular}%
}
\end{table}

\section{MetaResearcher Framework}
\label{sec:framework}

\begin{figure}[H]
    \centering
    \includegraphics[width=0.92\textwidth]{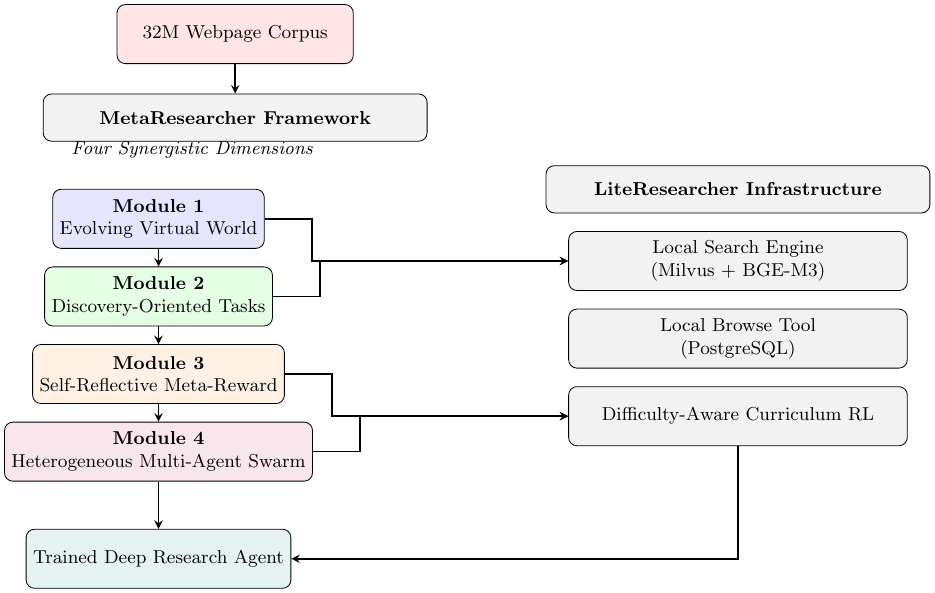}
    \caption{Conceptual overview of the MetaResearcher framework. The four innovation dimensions---Evolving Virtual World, Discovery-Oriented Tasks, Self-Reflective Meta-Reward, and Heterogeneous Multi-Agent Swarm---are integrated into a unified training pipeline built upon LiteResearcher's local search/browse environment.}
    \label{fig:overview}
\end{figure}

MetaResearcher extends the LiteResearcher infrastructure across four synergistic dimensions. Figure~\ref{fig:overview} provides a conceptual overview of the framework architecture.

\subsection{Evolving Virtual World}
\label{sec:evolving-world}

The first dimension addresses the fundamental limitation of static training environments. While LiteResearcher's virtual world of $\sim$32M webpages provides a realistic starting point, its frozen nature means that trained agents never encounter the temporal dynamics that characterize real-world research.

\subsubsection{Temporal Dynamics}

We introduce a \textit{time-evolving corpus} where the local web environment simulates information change over time. For each training episode, a random temporal context $t \in \mathcal{T}$ is sampled, and the corpus state reflects the information available at that simulated time point. This is implemented through:

\begin{itemize}
    \item \textbf{Versioned Documents:} Key documents (scientific papers, news articles) exist in multiple versions across the temporal dimension. For example, a paper's initial preprint may report finding X, a subsequent correction retracts X, and a later replication study confirms X with caveats.
    \item \textbf{Temporal Indexing:} The Milvus-based search engine~\cite{li2026literesearcher} indexes documents with temporal metadata. Search results are filtered and ranked according to the current simulated time, naturally exposing agents to information evolution.
    \item \textbf{Event Scripts:} Pre-scripted event sequences model real-world research dynamics, including paper retractions, press conference corrections, scientific consensus shifts, and emerging evidence that overturns prior conclusions.
\end{itemize}

\subsubsection{Adversarial Information Injection}

Beyond temporal dynamics, we introduce structured adversarial content designed to test and improve agents' epistemic resilience:

\begin{itemize}
    \item \textbf{High-Plausibility Misinformation:} Following the Synthetic Web methodology~\cite{shah2026synthetic}, we inject fabricated articles that mimic authoritative sources (prestigious journal formats, well-known author names) but contain factually incorrect claims.
    \item \textbf{Contradictory Expert Opinions:} We create clusters of seemingly credible sources that present conflicting conclusions on the same question, requiring agents to evaluate evidence quality rather than simply counting citations.
    \item \textbf{Subtle Manipulation:} More challenging variants include articles where only specific paragraphs contain misinformation while surrounding context is accurate, testing fine-grained source assessment.
\end{itemize}

The adversarial content is generated through a separate LLM pipeline with quality control, ensuring that misinformation is plausible enough to challenge the agent but distinguishable from ground truth through careful analysis. This approach transforms the training environment from a passive information store into an active teacher that systematically exposes weaknesses in the agent's research strategy.

\begin{figure}[H]
    \centering
    \includegraphics[width=0.9\textwidth]{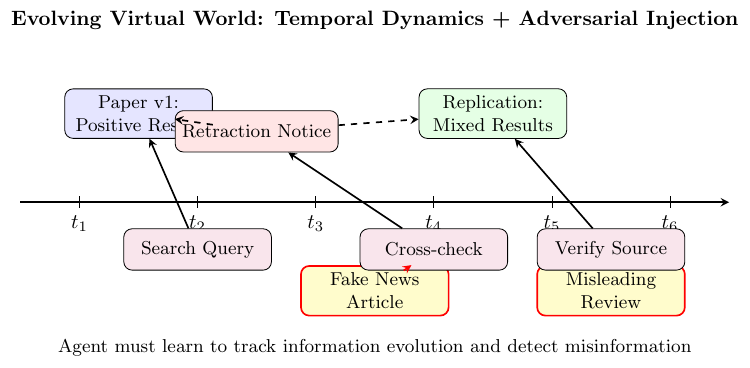}
    \caption{Evolving Virtual World mechanism. Documents evolve across a temporal axis with versioning (green/blue/red), while adversarial misinformation (yellow/red) is injected to test source credibility assessment. The agent must learn to track information evolution and detect deceptive content.}
    \label{fig:evolving}
\end{figure}

\subsection{Discovery-Oriented Tasks}
\label{sec:discovery-tasks}

The second dimension redefines the training task distribution to cultivate higher-order research capabilities beyond fact retrieval.

\subsubsection{Task Taxonomy}

We design three categories of discovery-oriented tasks, each targeting distinct cognitive capabilities:

\textbf{Hypothesis Generation (HG):} The agent is presented with documents from two seemingly unrelated research domains (e.g., ``coral reef bleaching patterns'' and ``financial market microstructure''). The task is to (a) retrieve and analyze relevant information from both domains, (b) identify structural or mechanistic parallels, and (c) formulate a novel, plausible hypothesis connecting the two domains. Evaluation is performed by an LLM judge assessing the hypothesis on novelty, plausibility, and evidence grounding.

\textbf{Contradiction Resolution (CR):} The agent encounters multiple sources that present conflicting accounts of the same phenomenon (e.g., three different explanations for a historical event, or competing theories about a scientific mechanism). The agent must (a) retrieve the full context of each position, (b) evaluate source credibility and evidence quality, (c) identify points of genuine disagreement versus semantic differences, and (d) produce a synthesized conclusion with explicit uncertainty quantification.

\textbf{Knowledge Gap Identification (KGI):} Given a well-studied research question, the agent must identify genuine gaps or inconsistencies in the available literature rather than simply summarizing known findings. This tests the agent's ability to recognize what information is \textit{missing}---a skill critical for genuine research but absent from existing benchmarks.

\begin{table}[H]
\centering
\caption{Discovery-oriented task taxonomy.}
\label{tab:tasks}
\renewcommand{\arraystretch}{1.15}
\resizebox{\linewidth}{!}{%
\begin{tabular}{lccc}
\toprule
\textbf{Task Type} & \textbf{Skill} & \textbf{Evaluation} & \textbf{Diff.} \\
\midrule
Hypothesis Gen. (HG) & Cross-domain synth. & Novelty, plausibility, grounding & H--E \\
Contradiction Res. (CR) & Source-critical reas. & Completeness, source weighing & M--H \\
Know. Gap Ident. (KGI) & Metacognitive aware. & Gap precision, coverage, action. & M--E \\
\bottomrule
\end{tabular}%
}
\end{table}

\subsubsection{Training Data Construction}

Following LiteResearcher's co-construction methodology~\cite{li2026literesearcher}, training data for these tasks is synthesized through a pipeline that combines:

\begin{enumerate}
    \item \textbf{Seed Document Sampling:} Documents are sampled from the local corpus based on topic diversity and information density metrics.
    \item \textbf{Task Generation:} An LLM-based generator produces task specifications (questions, expected research approach, and gold-standard evaluation criteria) from the seed documents.
    \item \textbf{Difficulty Calibration:} Each task is assigned a difficulty score based on the pass@8 performance of a reference model, following the curriculum RL paradigm~\cite{li2026literesearcher}.
    \item \textbf{Quality Filtering:} Tasks that are trivially solvable (pass@8 $>$ 0.9) or impossibly difficult (pass@8 $<$ 0.05) are filtered or redistributed across difficulty levels.
\end{enumerate}

\subsection{Self-Reflective Meta-Reward}
\label{sec:meta-reward}

The third dimension addresses the fundamental limitation of outcome-based reward by introducing a multi-component meta-reward function that jointly optimizes for correctness, efficiency, reflection quality, and behavioral diversity.

\begin{figure}[H]
    \centering
    \includegraphics[width=0.85\textwidth]{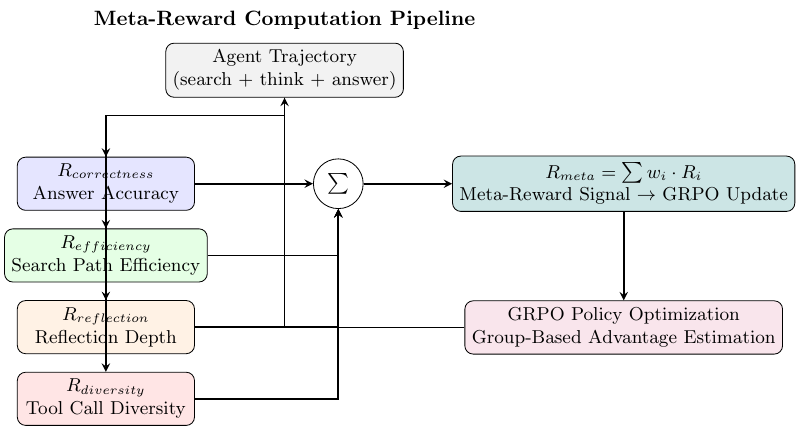}
    \caption{Self-Reflective Meta-Reward computation pipeline. The agent's trajectory is evaluated across four reward components, which are aggregated into a weighted meta-reward signal for GRPO-based policy optimization.}
    \label{fig:reward}
\end{figure}

\subsubsection{Reward Components}

The meta-reward $R_{\text{meta}}$ is defined as:

\begin{equation}
R_{\text{meta}} = w_c \cdot R_{\text{correctness}} + w_e \cdot R_{\text{efficiency}} + w_r \cdot R_{\text{reflection}} + w_d \cdot R_{\text{diversity}}
\label{eq:meta-reward}
\end{equation}

where $w_c, w_e, w_r, w_d$ are hyperparameters controlling the relative contribution of each component, with $\sum w_i = 1$.

\textbf{Correctness Reward ($R_{\text{correctness}}$):} Following LiteResearcher~\cite{li2026literesearcher} and the broader practice of outcome-based RL~\cite{gpt4o2025,qwen2025}, this is the binary outcome-based reward determined by an LLM judge comparing the agent's final answer against the reference answer. For GAIA-style tasks, the judge uses a correctness prompt; for Xbench-DS tasks, a Chinese-language evaluation prompt.

\textbf{Path Efficiency Reward ($R_{\text{efficiency}}$):} This reward component incentivizes information-economic search behavior:

\begin{equation}
R_{\text{efficiency}} = \begin{cases}
1.0 & \text{if agent finds answer in $\leq T_{\text{min}}$ steps} \\
1.0 - \alpha \cdot \frac{t - T_{\text{min}}}{T_{\text{max}} - T_{\text{min}}} & \text{if $T_{\text{min}} < t \leq T_{\text{max}}$} \\
0.0 & \text{if $t > T_{\text{max}}$}
\end{cases}
\label{eq:efficiency-reward}
\end{equation}

where $t$ is the number of tool calls, $T_{\text{min}}$ is the expected minimum steps for the task, $T_{\text{max}}$ is the maximum allowed steps, and $\alpha$ is a scaling factor.

\textbf{Reflection Depth Reward ($R_{\text{reflection}}$):} This is the key novel component that directly incentivizes self-reflective behavior within the agent's reasoning traces:

\begin{equation}
R_{\text{reflection}} = \beta_1 \cdot \mathbb{I}[\text{backtrack}] + \beta_2 \cdot \mathbb{I}[\text{strategy\_change}] + \beta_3 \cdot \frac{N_{\text{distinct\_sources}}}{N_{\text{total\_visits}}}
\label{eq:reflection-reward}
\end{equation}

where:
\begin{itemize}
    \item $\mathbb{I}[\text{backtrack}]$ is an indicator for explicit acknowledgments within \texttt{<think>} tags that a previous search direction was unproductive
    \item $\mathbb{I}[\text{strategy\_change}]$ detects changes in search strategy (e.g., switching from keyword search to author-based search after initial failure)
    \item The third term measures source diversity relative to total page visits, rewarding agents that explore genuinely different information sources rather than repeatedly visiting variants of the same source
\end{itemize}

\textbf{Tool Call Diversity Reward ($R_{\text{diversity}}$):} This component directly penalizes the repetitive action loop pathology:

\begin{equation}
R_{\text{diversity}} = \gamma \cdot \frac{|\text{unique\_queries}|}{|\text{total\_calls}|} + (1-\gamma) \cdot \frac{|\text{unique\_domains}|}{|\text{total\_visits}|}
\label{eq:diversity-reward}
\end{equation}

where the first term measures search query diversity and the second term measures webpage domain diversity. Both terms are normalized by the total number of calls/visits to prevent trivial maximization through extended search.

\begin{table}[H]
\centering
\caption{Meta-reward component design.}
\label{tab:reward}
\renewcommand{\arraystretch}{1.15}
\resizebox{\linewidth}{!}{%
\begin{tabular}{lccc}
\toprule
\textbf{Component} & \textbf{Formulation} & \textbf{Purpose} & \textbf{Behavior} \\
\midrule
$R_{\text{correct}}$ & Binary LLM-judge & Answer accuracy & Correct answering \\
$R_{\text{effic.}}$ & Step-count penalty & Concise search & Avoid excess calls \\
$R_{\text{reflect}}$ & Backtrack indicator & Self-correction & Error recovery \\
$R_{\text{divers.}}$ & Query/domain ratio & Exploration & Avoid repetition \\
\bottomrule
\end{tabular}%
}
\end{table}

\subsubsection{Optimization Procedure}

The meta-reward is integrated into the GRPO training loop following the on-policy, difficulty-aware curriculum paradigm established by LiteResearcher~\cite{li2026literesearcher}. For each training step:

\begin{enumerate}
    \item Sample a batch of tasks from the current curriculum difficulty level
    \item For each task, generate $G$ rollouts ($G=8$ following~\cite{li2026literesearcher})
    \item Compute $R_{\text{meta}}$ for each rollout trajectory
    \item Estimate group-based advantages: $A_i = \frac{R_{\text{meta}}^{(i)} - \mu_G}{\sigma_G}$
    \item Update policy via GRPO objective: $\mathcal{J}_{\text{GRPO}}(\theta) = \mathbb{E}\Bigl[\frac{1}{G}\sum_{i=1}^G \min\bigl(r_i(\theta) A_i,$\\$\qquad\qquad\qquad \text{clip}(r_i(\theta), 1-\epsilon, 1+\epsilon) A_i\bigr)\Bigr]$
\end{enumerate}

where $r_i(\theta) = \frac{\pi_\theta(o_i|q)}{\pi_{\theta_{\text{old}}}(o_i|q)}$ is the importance sampling ratio.

\subsection{Heterogeneous Multi-Agent Swarm}
\label{sec:multi-agent}

The fourth dimension transitions from a monolithic agent architecture to a distributed, specialized multi-agent system trained via coordinated reinforcement learning. Concurrent work such as DECOR~\cite{chen2026decor} has demonstrated the effectiveness of role decomposition in deep search through Planner/Filter/Answerer agents trained with hybrid rewards, while SAGE~\cite{peng2026sage} shows how multiple agents can co-evolve from a shared backbone through self-generated curricula. MetaResearcher extends these paradigms by introducing jointly-trained, lightweight specialized agents with a learned communication protocol optimized through coordinated GRPO.

\begin{figure}[H]
    \centering
    \includegraphics[width=0.85\textwidth]{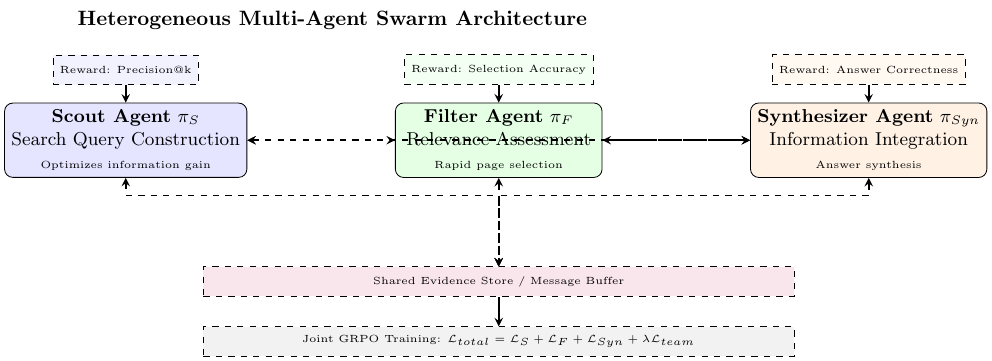}
    \caption{Heterogeneous Multi-Agent Swarm architecture. Three specialized agents---Scout, Filter, and Synthesizer---communicate through a shared evidence store and are jointly trained via GRPO with role-specific and team-level rewards.}
    \label{fig:swarm}
\end{figure}

\subsubsection{Agent Roles}

We define three specialized agent roles, each instantiated as a compact language model (4B parameters, matching the LiteResearcher scale):

\begin{itemize}
    \item \textbf{Scout Agent ($\pi_S$):} Specialized in search query construction. Given the current research question and the history of previous search attempts, the Scout generates optimized queries designed to maximize information gain. Its action space consists of search query strings formatted for the local Milvus search engine.

    \item \textbf{Filter Agent ($\pi_F$):} Specialized in relevance assessment. Given a set of search results (titles, snippets, URLs), the Filter rapidly identifies which pages are worth visiting for detailed reading. Its output is a ranked subset of URLs with estimated relevance scores and specific information goals for each selected page.

    \item \textbf{Synthesizer Agent ($\pi_{Syn}$):} Specialized in information integration. Given the collected search results and page contents, the Synthesizer produces the final answer by identifying patterns, resolving contradictions, and synthesizing fragmented information into coherent conclusions. It also manages the shared evidence store and decides when sufficient information has been gathered.
\end{itemize}

\begin{table}[H]
\centering
\caption{Multi-agent role definitions.}
\label{tab:roles}
\renewcommand{\arraystretch}{1.15}
\resizebox{\linewidth}{!}{%
\begin{tabular}{lccc}
\toprule
\textbf{Agent} & \textbf{Input} & \textbf{Output} & \textbf{Reward} \\
\midrule
Scout $\pi_S$ & Question + history & Optimized queries & Precision@k \\
Filter $\pi_F$ & Results + snippets & Ranked URLs & Selection F1 \\
Synthes. $\pi_{Syn}$ & Page contents & Synthesized answer & Correctness \\
\cmidrule{2-4}
\multicolumn{1}{l}{\textbf{Shared}} & --- & --- & Task success \\
\bottomrule
\end{tabular}%
}
\end{table}

\subsubsection{Communication Protocol}

The three agents communicate through a structured message passing interface, implemented as shared memory within LiteResearcher's existing message framework:

\begin{equation}
m_{t}^{(i \rightarrow j)} = f_{\text{comm}}(\text{state}_t^{(i)}, g_{t}^{(i)}, h_t)
\label{eq:comm}
\end{equation}

where $m_{t}^{(i \rightarrow j)}$ is the message from agent $i$ to agent $j$ at time $t$, $f_{\text{comm}}$ is a learned communication function, $\text{state}_t^{(i)}$ is agent $i$'s internal state, $g_{t}^{(i)}$ is its current goal, and $h_t$ is the shared interaction history.

Rather than pre-specifying the communication format, we allow agents to develop an emergent communication protocol through the RL training process, where the content and format of inter-agent messages are optimized for team-level task performance.

\subsubsection{Training Framework}

All three agents share the same base architecture (transformer decoder, matching LiteResearcher-4B) but are trained with distinct reward signals and data distributions:

\begin{equation}
\mathcal{L}_{\text{total}} = \mathcal{L}_{\text{GRPO}}(\pi_S) + \mathcal{L}_{\text{GRPO}}(\pi_F) + \mathcal{L}_{\text{GRPO}}(\pi_{Syn}) + \lambda \cdot \mathcal{L}_{\text{team}}
\label{eq:multi-agent-loss}
\end{equation}

where $\mathcal{L}_{\text{GRPO}}(\pi_i)$ is the individual GRPO loss for each agent with role-specific rewards, $\mathcal{L}_{\text{team}}$ is a shared team-level reward that encourages effective coordination, and $\lambda$ controls the balance between individual specialization and team coherence.

The individual reward for the Scout prioritizes search result quality (precision@k of results returned given the query), the Filter's reward prioritizes selection accuracy (whether recommended pages contain task-relevant information), and the Synthesizer's reward is primarily based on final answer correctness. The team reward component provides a shared signal proportional to overall task success, coordinating the three agents toward a common objective. This role-specific reward design is complementary to offline RL approaches such as Retrospex~\cite{retrospex2025}, which uses offline-trained critics for action rescoring in agentic tasks.

\section{Experimental Design}
\label{sec:experiments}

In this section, we describe the planned experimental validation of MetaResearcher. The experiments are designed to evaluate each innovation dimension both independently and in combination, with ablation studies identifying marginal contributions.

\subsection{Evaluation Benchmarks}

We evaluate MetaResearcher on three categories of benchmarks:

\begin{enumerate}
    \item \textbf{Standard Benchmarks:} GAIA (all levels) and Xbench-DS, following the LiteResearcher evaluation protocol~\cite{li2026literesearcher}. These measure the framework's ability to maintain or improve performance on existing metrics.

    \item \textbf{Epistemic Robustness Benchmark:} A new benchmark suite comprising questions where optimal performance requires (a) detecting and discounting misinformation, (b) resolving contradictory sources, and (c) tracking information evolution over time. This benchmark directly tests the Evolving Virtual World's training effects.

    \item \textbf{Discovery Task Benchmark:} A set of hypothesis generation and contradiction resolution tasks with human-annotated gold standards, designed to measure the framework's success in cultivating higher-order research capabilities.
\end{enumerate}

\subsection{Ablation Studies}

We conduct comprehensive ablation studies:

\begin{itemize}
    \item \textbf{Environment Ablation:} Compare fully dynamic environment vs. static (original LiteResearcher) vs. dynamic without adversarial injection
    \item \textbf{Reward Ablation:} Compare full meta-reward vs. outcome-only vs. outcome + efficiency vs. outcome + reflection depth
    \item \textbf{Architecture Ablation:} Compare full multi-agent swarm vs. single-agent with same total parameters vs. multi-agent without shared team reward
    \item \textbf{Task Ablation:} Compare training with and without discovery-oriented tasks, measuring transfer effects on standard benchmarks
\end{itemize}

\subsection{Implementation Details}

The framework is implemented as an extension of the LiteResearcher codebase~\cite{li2026literesearcher}. Key implementation details:

\begin{itemize}
    \item Base model: Qwen2.5-4B-Instruct~\cite{qwen2025}, following~\cite{li2026literesearcher}
    \item Local search engine: Milvus with BGE-M3 embeddings
    \item Local browse tool: PostgreSQL-based page store
    \item Multi-agent communication: Shared message buffer with learned gating
    \item Curriculum stages: 4-stage difficulty progression (Easy $\rightarrow$ Medium $\rightarrow$ Hard $\rightarrow$ Expert)
    \item Training hardware: 8 $\times$ A100-80GB GPUs
    \item Estimated training budget: $\sim$2000 GPU-hours for full pipeline
    \item Synthetic data generation following agentic LLM training paradigms~\cite{llamaagent2025}
\end{itemize}

\subsection{Hypotheses}

We formalize the following testable hypotheses, each targeting a specific innovation dimension:

\begin{table}[H]
\centering
\caption{Experimental hypotheses.}
\label{tab:hypotheses}
\footnotesize
\renewcommand{\arraystretch}{1.1}
\begin{tabular}{c p{2cm} p{2.2cm} p{2cm} p{2cm}}
\toprule
\textbf{H} & \textbf{Dimension} & \textbf{Expected Outcome} & \textbf{Metric} & \textbf{Baseline} \\
\midrule
H1 & All (integ.) & GAIA $\geq$73.0\% (+1.7) & GAIA all levels & LR 71.3\% \\
H2 & Evolving W. & Robustness $\uparrow$20\% & Epistemic bench & Static env. \\
H3 & Meta-Reward & Loops $\downarrow$50\% & Identical calls & Outcome RL \\
H4 & Swarm & Sum $>$ indiv. & GAIA+Xbench & Single agent \\
H5 & Discovery & No degrad. & GAIA transfer & Pre-Disc. tr. \\
\bottomrule
\end{tabular}
\end{table}

\section{Discussion}
\label{sec:discussion}

\subsection{Implications}

MetaResearcher's four-dimensional approach to scaling deep research agents has several implications for the broader field. First, the integration of adversarial elements during training---rather than only during evaluation---represents a paradigm shift in how we think about agent robustness. Current systems treat epistemic vulnerabilities as evaluation artifacts to be measured; we argue they should be treated as training targets to be optimized.

Second, the transition from outcome-based to process-based rewards addresses a fundamental limitation of current RL-for-agent pipelines. By explicitly rewarding efficient search paths, self-reflective behavior, and tool use diversity, we create training signals that more closely align with genuine research quality rather than superficial answer correctness.

Third, the heterogeneous multi-agent architecture aligns with the growing recognition that specialized, modular systems---rather than monolithic models---represent the most scalable path toward capable AI agents. This echoes findings from distributed systems and organizational theory, where specialization and coordination consistently outperform generalist approaches on complex tasks.

\subsection{Limitations}

Several limitations warrant acknowledgment. The adversarial content generation pipeline itself requires careful quality control to ensure that injected misinformation is challenging but detectable. Poorly calibrated adversarial content could either be trivially identifiable (providing no training signal) or unfairly deceptive (teaching agents to be overly skeptical of legitimate information). We address this through multi-stage human review of adversarial content quality.

The multi-agent training framework introduces additional computational overhead for inter-agent communication and synchronized training. While we expect the performance gains to justify this cost, the practical trade-off between specialization benefits and coordination overhead requires empirical quantification.

Finally, while discovery-oriented tasks push beyond fact retrieval, they introduce evaluation challenges. Hypothesis generation, in particular, lacks clear ``correct/incorrect'' boundaries, and our LLM-judge-based evaluation may not fully capture the quality of generated hypotheses.

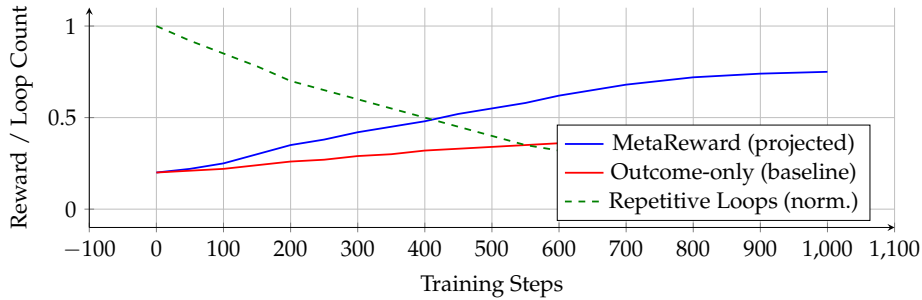
\begin{figure}[H]
    \centering
    \begin{tikzpicture}[scale=0.85]
    \begin{axis}[
        xlabel={Training Steps}, ylabel={Reward / Loop Count},
        width=0.85\textwidth, height=5cm,
        legend pos=south east,
        xmin=0, xmax=1000,
        ymin=0, ymax=1,
        grid=both, minor grid style={gray!20},
        axis lines=left, enlargelimits=true
    ]
    \addplot[blue, thick] coordinates {
        (0,0.20) (50,0.22) (100,0.25) (150,0.30) (200,0.35)
        (250,0.38) (300,0.42) (350,0.45) (400,0.48) (450,0.52)
        (500,0.55) (550,0.58) (600,0.62) (650,0.65) (700,0.68)
        (750,0.70) (800,0.72) (850,0.73) (900,0.74) (950,0.745) (1000,0.75)
    };
    \addlegendentry{MetaReward (projected)}

    \addplot[red, thick] coordinates {
        (0,0.20) (50,0.21) (100,0.22) (150,0.24) (200,0.26)
        (250,0.27) (300,0.29) (350,0.30) (400,0.32) (450,0.33)
        (500,0.34) (550,0.35) (600,0.36) (650,0.37) (700,0.38)
        (750,0.39) (800,0.395) (850,0.40) (900,0.405) (950,0.41) (1000,0.41)
    };
    \addlegendentry{Outcome-only (baseline)}

    \addplot[green!50!black, dashed, thick] coordinates {
        (0,1.0) (50,0.92) (100,0.85) (150,0.78) (200,0.70)
        (250,0.65) (300,0.60) (350,0.55) (400,0.50) (450,0.45)
        (500,0.40) (550,0.35) (600,0.32) (650,0.28) (700,0.25)
        (750,0.22) (800,0.20) (850,0.18) (900,0.17) (950,0.16) (1000,0.15)
    };
    \addlegendentry{Repetitive Loops (norm.)}
    \end{axis}
    \end{tikzpicture}
    \caption{Projected training dynamics. The meta-reward trajectory (blue) shows accelerated improvement over outcome-only baseline (red), while normalized repetitive action loops (green, dashed) exhibit a projected decline of $>$50\%.}
    \label{fig:training-dynamics}
\end{figure}

\section{Conclusions}
\label{sec:conclusion}

We have presented MetaResearcher, a comprehensive framework for scaling deep research agent training across four synergistic dimensions: an Evolving Virtual World that injects temporal dynamics and adversarial information into the training environment, Discovery-Oriented Tasks that transcend simple fact retrieval, a Self-Reflective Meta-Reward mechanism that jointly optimizes for correctness, efficiency, reflection quality, and behavioral diversity, and a Heterogeneous Multi-Agent Swarm architecture comprising specialized Scout, Filter, and Synthesizer models.

Built upon the LiteResearcher infrastructure, MetaResearcher inherits its zero-marginal-cost training paradigm while systematically addressing its four fundamental limitations: static environments, fact-retrieval-only tasks, outcome-only rewards, and monolithic architecture. The framework targets substantial improvements in both standard benchmark performance and epistemic robustness under adversarial conditions.

The codebase, trained models, and training data will be released as open-source extensions of the LiteResearcher project, enabling reproducible research and community-driven advancement. We believe the four-dimensional approach outlined in this work represents a meaningful step toward deep research agents that can serve as genuine collaborators in the scientific discovery process.

\textbf{Author Contributions:} Conceptualization, W.Y. and S.L.; methodology, W.Y., M.Y., and S.L.; software, Z.Z. and H.D.; validation, J.W., B.L., and S.L.; investigation, W.Y. and M.Y.; resources, S.L.; data curation, M.Y. and J.W.; writing---original draft preparation, W.Y. and S.L.; writing---review and editing, S.L. and J.W.; visualization, W.Y. and Z.Z.; supervision, S.L.; project administration, S.L.; funding acquisition, S.L. All authors have read and agreed to the published version of the manuscript.

\textbf{Funding:} This research received no external funding.

\textbf{Data Availability Statement:} The LiteResearcher-Data dataset is publicly available on HuggingFace. All code and trained models will be released upon publication.

\textbf{Conflicts of Interest:} The authors declare no conflicts of interest.

\textbf{AI Usage Disclosure:} Portions of this manuscript were drafted with the assistance of large language models. All content was reviewed, edited, and approved by the human authors, who take full responsibility for the intellectual content and accuracy of the work.

\end{document}